\pdfoutput=1

\documentclass[11pt]{article}

\usepackage{ACL2023}

\usepackage{times}
\usepackage{latexsym}

\usepackage{graphicx}
\usepackage{enumitem}
\usepackage{amsmath, amsfonts, amssymb}

\usepackage[T1]{fontenc}

\usepackage[utf8]{inputenc}

\usepackage{microtype}

\usepackage{inconsolata}

\title{Towards Uncovering How Large Language Model Works: \\ An Explainability Perspective}

\author{Haiyan Zhao\textsuperscript{1}, Fan Yang\textsuperscript{2}, Bo Shen\textsuperscript{1}, Himabindu Lakkaraju\textsuperscript{3}, Mengnan Du\textsuperscript{1},\\
  \textsuperscript{1}New Jersey Institute of Technology \,
  \textsuperscript{2}Wake Forest University \,
  \textsuperscript{3}Harvard University\\
  \small\texttt{\{hz54,bo.shen,mengnan.du\}@njit.edu},\, \small\texttt{yangfan@wfu.edu},\, \small\texttt{hlakkaraju@hbs.edu}
}

\begin{document}
\maketitle
\begin{abstract}
Large language models (LLMs) have led to breakthroughs in language tasks, yet the internal mechanisms that enable their remarkable generalization and reasoning abilities remain opaque. This lack of transparency presents challenges such as hallucinations, toxicity, and misalignment with human values, hindering the safe and beneficial deployment of LLMs. This paper aims to uncover the mechanisms underlying LLM functionality through the lens of explainability. First, we review how knowledge is architecturally composed within LLMs and encoded in their internal parameters via mechanistic interpretability techniques. Then, we summarize how knowledge is embedded in LLM representations by leveraging probing techniques and representation engineering. Additionally, we investigate the training dynamics through a mechanistic perspective to explain phenomena such as grokking and memorization. Lastly, we explore how the insights gained from these explanations can enhance LLM performance through model editing, improve efficiency through pruning, and better align with human values.
\end{abstract}

\section{Introduction}
Large language models (LLMs) such as GPT-4~\citep{openai2023gpt4}, LLaMA-2~\citep{touvron2023llama2}, Claude-3~\citep{AnthropicAI2023}, and Gemini~\citep{team2023gemini} have led to tremendous advancements in language understanding and generation, achieving state-of-the-art performance in a wide array of real-world tasks. Despite their superior performance across various tasks, the ``how'' and ``why'' behind their generalization and reasoning abilities are still not well understood. This lack of understanding poses several challenges. First, LLMs frequently generate hallucinations and factually incorrect output, which complicates efforts to improve their performance. Second, as LLMs become more powerful, problems surrounding potential toxicity, unfairness, and dishonesty threaten to undermine user trust. Third, the impressive generalization capabilities of LLMs suggest that they may be acquiring and leveraging knowledge in ways that fundamentally differ from traditional machine learning approaches. Therefore, there is an urgent need to delve deeper into the inner workings of LLMs to fully address these issues. Gaining insights into how these models operate is a crucial step towards developing robust safeguards and ensuring their responsible deployment.

\begin{figure*}[t]
    \centering
    \includegraphics[width=0.95\textwidth]{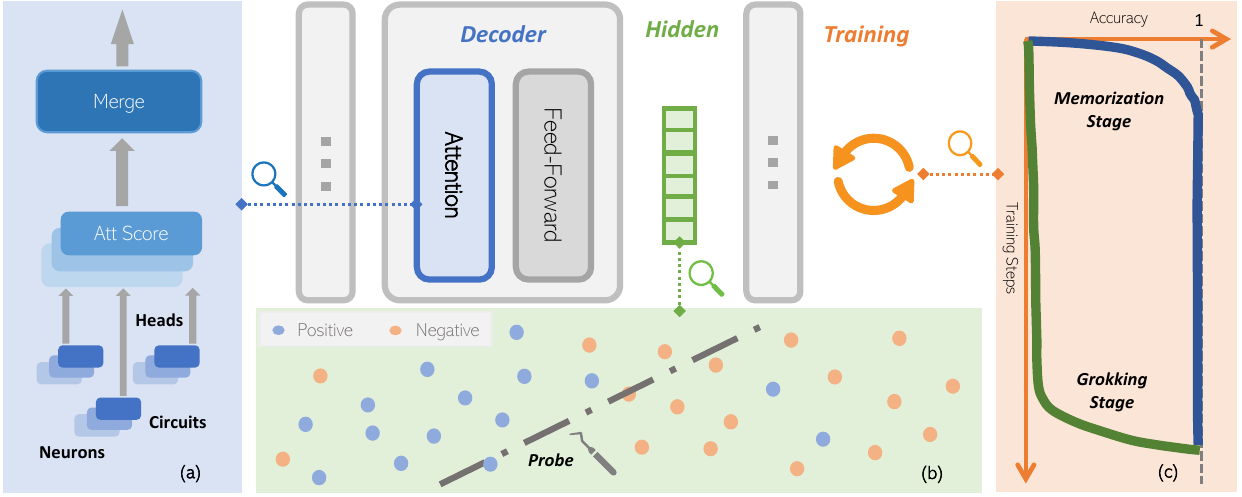}
    \caption{In this work, we review existing progress on how LLMs work, including: a) how knowledge is architecturally composed within model components; b) what knowledge is encoded in intermediate representations; and c) how generalization abilities are achieved during the training process.}
    \label{fig:overall}
\end{figure*}

In this paper, we provide a systematic overview of the existing literature that uncovers the working mechanisms of LLMs using existing explainability techniques (Figure~\ref{fig:overall}). First, we provide a summary of findings on probing trained LLMs to understand how knowledge is composed in model architectures. To achieve this, the main explainability technique used is mechanistic interpretability. It focuses on the functionality of each model component and interprets how models operate at the level of neurons, circuits, and attention heads. Second, we examine how knowledge is encoded internally in intermediate representations. To this end, representation engineering is adopted to explain specific behavior of the model, such as dishonesty, by analyzing hidden representations~\citep{zou2023representation}. Third, we inspect the model training process to understand the development of generalization abilities during the training process.
Finally, we review how insights from the aforementioned analysis help us improve models in terms of higher performance through model editing, better efficiency through pruning, and better human alignment.

Our work differs from existing survey articles on the explainability of LLMs~\citep{zhao2023explainability,wu2024usable,luo2024understanding}, which either summarize explainability techniques or discuss their utilities. On the contrary, our goal is to review existing studies to uncover how LLMs function and identify the factors that contribute to their reasoning abilities via using explainability techniques and monitoring the training process. We review the state-of-the-art insights on the inner workings of LLMs and explore how these insights can further enhance model performance and benefit humans. We believe that a deep understanding of how LLMs work is crucial for their safe and beneficial deployment in real-world applications.

\section{How is Knowledge Composed in Model Architectures?}
LLMs are built on extensive training datasets and intricate model architectures, which contribute to their remarkable emergent abilities~\citep{wei2022emergent}. However, the exact mechanisms through which these models acquire and process vast amounts of knowledge remain unclear. Additionally, the contributions of individual model components to the overall function have been largely unexplored. To fully understand LLMs, recent studies have shifted to make use of mechanistic interpretability (more details are given in Section~\ref{sec:appendix-mi} at the Appendix) to reverse engineer LLMs at a more granular level such as neurons, attention heads, and connections.

\subsection{Neurons}
Neurons serve as the basic units for storing knowledge and patterns within LLMs. They are observed to be \emph{polysemantic}, meaning that an individual neuron can be activated on multiple unrelated terms~\citep{olah2020zoom,bills2023language}. This characteristic presents a significant challenge in mechanistically understanding how models operate. Despite the challenge, recent work has explored the underlying causes of this polysemantic nature. Two key concepts have emerged as instrumental in unraveling its formation: \emph{Superposition}~\citep{olah2020zoom} and \emph{Monosemanticity}~\citep{bricken2023monosemanticity}.  

\subsubsection{Superposition}
Superposition describes the phenomenon where a feature can be spread across multiple neurons, meanwhile a neuron can also be mixed up with multiple features. Some researchers believe that this mechanism originated from an excessive number of features compared to the number of neurons~\citep{olah2020zoom,elhage2022superposition}. 
In exploring this concept through a toy example, i.e. a ReLU network, researchers have found that superposition allows for the representation of additional features. However, to mitigate interference, a nonlinear filter needs to be introduced~\citep{elhage2022superposition}. When features are sparse, superposition effectively supports the representation of these features and allows computations such as the absolute value function~\citep{elhage2022superposition}. Neurons within models can be either monosemantic or polysemantic.

Others argue that polysemanticity arises incidentally due to factors encountered during the training process such as regularization and neural noise~\citep{lecomte2024causes}. Mathematical demonstrations have shown that a constant fraction of feature collisions, introduced through random initialization, can always result in polysemantic neurons, even when the number of neurons exceeds the number of features~\citep{lecomte2024causes}. 

Another study investigates polysemanticity through the lens of the ``feature capacity'', denoting the fraction of embedding dimensions consumed by a feature in the representation space~\citep{scherlis2022polysemanticity}. By analyzing one-layer and two-layer toy models, this work indicates that features are represented based on their importance in reducing loss. More important features are allocated their own dimensions, while the less critical ones may be overlooked, and the rest will share embedding dimensions~\citep{scherlis2022polysemanticity}. Features only end up sharing dimensions when assigning additional capacity will not result in loss decreasing~\citep{scherlis2022polysemanticity}. Moreover, the relationship between superposition and feature importance has been demonstrated on LLMs~\citep{gurnee2023finding}. Experiments show that the early layers tend to represent many features in superposition, while the middle layers include dedicated neurons to represent high-level features~\citep{gurnee2023finding}.

\subsubsection{Monosematicity}
Monosemantic neurons, associated with a single concept, are much easier to interpret than polysemantic neurons. Investigating the factors that enhance monosemanticity is meaningful to model interpretation. A research using toy models reveals that changing the loss minimum could improve monosemanticity. Such loss minimum usually coexists with negative biases~\citep{jermyn2022engineering}. However, in reality building a purely monosemantic model is infeasible due to the unmanageable loss~\citep{bricken2023monosemanticity}. Another line of studies seeks to disentangle superposition to reach a monosemantic understanding. The spare autoencoder emerges as a promising tool for this purpose, particularly through the method dictionary learning where features are predefined~\citep{sharkeyInterimResearchReport}. The effectiveness of this approach largely depends on the comprehensiveness of the pre-defined dictionary. \citet{bricken2023monosemanticity} using a one-layer transformer model with a 512-neuron MLP layer highlights this approach. The sparse autoencoder is trained on MLP activations from 8B data points, with autoencoder sizes ranging from 512 to 13,100 features. Larger autoencoders are able to achieve finer granularity in interpreting features, revealing details that cannot be discovered at the neuron level. These identified features can be used to manipulate the model's output, offering new ways to control and understand LLMs~\citep{bricken2023monosemanticity}.

\subsection{Circuits}\label{sec:circuits}
\begin{figure}[t]
    \centering
    \includegraphics[width=0.39\textwidth]{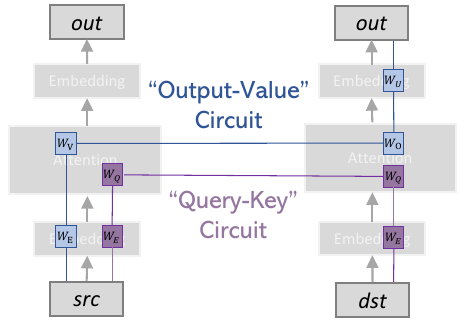}
    \caption{An illustration of a Transformer circuit, which is a key concept in mechanistic interpretability.}
    \label{fig:circuit}
\end{figure}

Circuit is one of the core concepts in the field of mechanistic interpretability (see Figure~\ref{fig:circuit}). It was originally proposed to reverse engineer vision models, in which individual neurons and their connections are viewed as functional units~\citep{olah2020zoom}. Researchers have found that features in former layers of models act as fundamental units, such as edge detectors. These features are combined through weights to form a circuit unit. This viewpoint is partially evidenced by a few understandable neuron units (or circuits) performing specific functions, such as curve detectors~\citep{cammarata2020curve} and high-low frequency detectors~\citep{schubert2021highlow}. Several interesting phenomena have been observed in these circuits. For example, symmetric transformations of basic features, including copying, scaling, flipping, coloring, rotating, can be achieved with basic neurons known as ``equivariance'' or ``motif''~\citep{olah2020naturally}.

Despite rich insights from vision models, transformer models, with their unique architecture featuring attention blocks, present new challenges. To address these, a mathematical framework specifically for \textit{transformer circuits} has also been proposed~\citep{elhage2021mathematical}. This framework simplifies the complex architecture of LLMs by focusing on decoder-only transformer models that have no more than two layers, all made up entirely of attention blocks. Within this toy model, the transformer encompasses input embedding, residual stream, attention layers, and output embeddings. Attention layers read information from the residual stream and then write their output back into it. Consequently, communication is achieved through read and write operations at the layer level.

Each attention head works independently and in parallel, contributing its output to the residual stream. These heads consist of key, query, output, and value vectors, represented as $W_K, W_Q, W_O, \text{and} W_V$. There are two types of circuits: i) ``query-key'' (QK) circuits; ii) ``output-value'' (OV) circuits~\citep{elhage2021mathematical}, as shown in Figure \ref{fig:circuit}. The QK circuits, formed by $W_Q^T W_K$, play a crucial role in determining which previously learned token to copy information from~\citep{elhage2021mathematical}. It is essential for models to recall and retrieve information from earlier context. 
Conversely, the OV circuits, composed of $W_O W_V$, determine how the current token influences the output logits~\citep{elhage2021mathematical}. 

The result shows that transformers with no layer can model bigram statistics, predicting the next token from the source token. Adding one layer allows the model to capture both bigram and ``skip-trigram'' patterns. Interestingly, with two layers, transformer models give rise to a concept termed as ``\textit{induction head}'' (Section \ref{sec:atten-head} ). These induction heads exist in the second layer and beyond. Usually, they are composed of heads from their previous layer, which is useful in suggesting the next token based on the present ones~\citep{elhage2021mathematical}.

\subsection{Attention heads}\label{sec:atten-head}
A special type of attention head called \emph{induction head} is assumed critical in enabling in-context learning abilities within LLMs~\citep{brown2020language}, due to the co-occurrence of induction heads and in-context learning~\citep{olsson2022context}. Induction heads also refer to a kind of circuits that complete the pattern by prefix matching and copying previously occurred sequences~\citep{olsson2022context}. They are composed of two heads: the first attention head is from the previous layer attending to previous tokens that are followed by the current token, which achieves prefix matching and provides the attend-to token(the token following current token). The second head, i.e. induction head, copies the attend-to token and increases its output logits. More specifically, this rule means that if models have seen similar patterns such as ``[A*][B*]'' given current token ``[A]'', these models are able to predict ``[B]''~\citep{olsson2022context}. Despite the single token used in the toy example, long prefix matching such as three consecutive tokens has also been observed in related work~\citep{chan2022causal}.

As a result, layers with induction heads possess more powerful in-context learning abilities than simple copying. In addition, multiple empirical studies have demonstrated the causal relationships between induction heads and in-context learning abilities by observing the change of in-context learning abilities after manipulating induction heads~\citep{olsson2022context, chan2022causal}. Although this theory offers a comprehensive explanation of the mechanisms behind transformer models with only two attention layers, further ablation studies are still needed to validate its effectiveness. It is also important to note that this framework is exclusively based on attention heads, without incorporating MLP layers.

\section{What Knowledge is Encoded in Intermediate Representations?}

In the previous section, we summarize existing studies on the architectural composition of knowledge within LLMs, with a focus on their structural components. We highlight how components of LLMs function differently. In this section, we introduce an in-depth review of the knowledge encoded by \emph{representations of LLMs}, including world knowledge and factual knowledge captured within these models. We examine how factors such as the depth of layers and the scale of models influence this encoding process.

\subsection{Probing World and Factual Knowledge}
To investigate whether the representations of LLMs encode world knowledge and factual knowledge, probing techniques offer crucial insights into the structure and dynamics of these representations. Specifically, probing techniques can identify specific directions within the representation space, and these directions are essential for understanding certain behaviors and the encoding of knowledge~\citep{zou2023representation,liu2023aligning}. 

Recent studies have demonstrated that LLMs can learn world models and encode them in their representations for certain tasks. One study successfully uses a set of non-linear probes to uncover world representations within models, specifically in the context of the game of Othello~\citep{li2022emergent}. It demonstrates models' ability to track the board state, and make predictions without being explicitly to do so~\citep{li2022emergent}. Furthermore, another work finds that linear representation structures can also perform well on predictions, simply by altering the expression of the board state at each timestamp~\citep{nanda2023emergent}. The linear and non-linear explanations reveal how models perceive the world naturally, which might be different from humans. Additionally, by analyzing representations of spatial datasets, one study reveals the model's ability to learn linear representations of space and time across multiple levels~\citep{gurnee2023language}.

LLMs are also capable of encoding factual knowledge.~\citet{marks2023geometry} craft self-curated true/false datasets to study the geometry of representations of true/false statements derived from a model's residual stream. By applying principal component analysis (PCA), a clear linear structure emerges. The truth directions are leveraged to mediate the model's dishonest behaviors locally. Another research avenue explores vectors related to toxicity within MLP blocks through singular value decomposition (SVD). The identified dimensions are simply subtracted to efficiently achieve mitigation~\citep{lee2024mechanistic}.

Function vectors have also been discovered within the attention heads of LLMs, which trigger the execution of a certain task across diverse inputs. For example, ~\citet{todd2023function} found that these function vectors are shown in various in-context learning tasks, and can execute related tasks despite zero-shot inputs. Also, causal interventions at the neuron level can help identify the individual neurons encoding spatial coordinates and time information~\citep{gurnee2023language}.

Lastly, representations associated with undesirable behaviors of LLMs, such as dishonesty, toxicity, hallucinations, can also be extracted. Typically, a direction in the representation space is identified as contributing to a specific behavior. This direction will then be used to adjust the representations so that models' behaviors can be controlled~\citep{zou2023representation}. For example, ~\citet{li2024inference} employs this technique to probe and enhance the truthfulness of models.~\citet{azaria2023internal} also successfully distinguishes the truthfulness of statements by simply training a classifier on model representations. A recent work has been developed to identify hallucination tokens from the response by integrating a range of classifiers that are trained on each layer from separate hidden parts: MLPs and attention layers~\citep{ch2023androids}. 

\subsection{Role of Layer Depth and Model Scale}
The influence of layer depth and model scale on representations has been an interesting research direction. Empirically, research shows that a range of knowledge is well trained until the middle layers. For example, ~\citet{gurnee2023language} demonstrate that space and time representations reach the best quality up to half of the layers in a range of open-source LLMs. 
Besides, the function vectors with strong causal effects are also collected from the middle layers of LLMs, while the effects are near zero in the deeper layers~\citep{todd2023function}. Furthermore, another study shows that different levels of concepts are well learned in different layers, where simpler tasks are learned in the early layers while complex tasks can only be well learned in the deeper layers~\citep{jin2024exploring,ju2024large}. However, the underlying reason why the middle layers perform so well remains unexplored. 

It is generally believed that more capabilities are gained as the models scale up~\citep{wei2022emergent}. Some recent studies have also supported this hypothesis in certain cases. For example, the space and time representations are more precise as the models scale up~\citep{gurnee2023language}. But the inner mechanism leading to better performance when model scales remain unknown.

\section{How is Generalization Ability Achieved During Training Process?}

In the preceding sections, we analyze LLMs in a post-hoc manner, focusing on neurons, connections, attention heads, and representations to understand how knowledge is acquired within models. In this section, we discuss the dynamic training process of models to understand how the generalization ability is achieved during the training process. We will particularly examine two important phenomena observed in relation to generalization: grokking and memorization. Here, grokking indicates the phenomenon where models suddenly improve validation accuracy after overfitting. Investigating grokking can shed light on how generalization emerges during training. Moreover, examining memorization, where models rely on statistical patterns rather than causal relationships, can help disentangle the roles of generalization versus the roles of memorization in model behaviors.

\subsection{Understanding Grokking}\label{sec:understanding-grokking}
\textit{Grokking} is a phenomenon in which models suddenly improve their validation accuracy after severely overfitting on over-parameterized neural networks~\citep{power2022grokking}. The surge in validation accuracy is generally interpreted as a gain of generalization ability.

\subsubsection{A Data Perspective}
Experiments implemented on a two-layer decoder-only transformer network have shown that grokking is closely related to factors such as data, representations, and regularization. Smaller datasets require more optimization steps for grokking to occur~\citep{power2022grokking}. Conversely, more samples can decrease the number of steps needed for generalization~\citep{zhu2024critical}. The minimal amount of data needed for grokking also depends on the minimal number of data points required to learn a robust representation~\citep{liu2022towards}. Furthermore, it has been found that generalization often coincides with well-structured embeddings. Additionally, regularization measures can accelerate the onset of grokking, with weight decay standing out as particularly effective in strengthening generalization capabilities~\citep{liu2022towards}. A recent study proposes that massive datasets in LLMs make grokking less conceivable~\citep{zhu2024critical}.

\subsubsection{Weight Norms}
When examining the weight norms of the final layers in models that do not use regularization techniques, a phenomenon, termed as \textit{slingshot mechanism}, has been observed. It describes a cyclic behavior during the terminal phase of training, where there are oscillations between stable and unstable regimes, i.e., training loss spike. The spike co-occurs with a phase where weight norms grow, followed by a phase of norm plateau.~\citet{thilak2022the} point out that grokking, non-trivial feature adaptation, occurs only at the beginning of slingshots. The appearance of the slingshot effect and grokking can be modulated by adjusting the optimizer parameters, especially when using adaptive optimizers such as Adam~\citep{kingma2014adam}. However, it is unclear whether this observation holds universally across various scenarios. 

Additionally, another concept called the \textit{LU mechanism} has also been proposed, describing dynamics between loss and weight norms~\citep{liu2022omnigrok}. In algorithmic datasets, an L-shaped training loss and a U-shaped test loss reduction concerning weight norms are identified, implying an optimal range for initializing weight norms. Nevertheless, this finding does not seamlessly transfer to real-world machine learning tasks, where large initialization and small weight decay are often necessary.~\citet{lyu2023dichotomy,mohamadi2023grokking} attribute it to a competition between the early-phase implicit bias favoring kernel predictors induced by large initialization and a late-phase implicit bias favoring min-norm/margin predictors promoted by small weight decay. Similarly, \citet{merrill2023tale} conclude that this competition manifests a competition between a dense subnetwork in the initial phase and a sparse one after grokking.

\subsubsection{Test Loss}
\textit{Double descent} captures the pattern where a model's test accuracy at the log level initially improves, then drops due to overfitting, and finally increases again after gaining generalization abilities~\citep{nakkiran2021deep}. This pattern is more noticeable in the test loss. A unified framework has been developed to integrate grokking with double descent, treating them as two manifestations of the same underlying process~\citep{davies2023unifying}. The framework attributes the transition of generalization to slower pattern learning, which has been further supported by~\citet{kumar2023grokking}. This transition is demonstrated to exist at the level of both epochs and models.

\subsection{Memorization}
\textit{Memorization} often refers to the phenomenon that models predict with statistical features rather than causal relations. The study using slightly corrupted algorithmic datasets with two-layer neural models has revealed that memorization can coexist with generalization. And memorization can be mitigated by pruning relevant neurons or by regularization~\citep{doshi2023grok}. Although different regularization methods might not share learning goals, they all contribute to better representations. And the training process in the study consists of two stages: i) the grokking process, ii) the decay of memorization learning~\citep{doshi2023grok}. However, the underlying causes behind this process are not yet fully understood. Besides, the assumption that regularization is the key to this process is under debate, especially in light of observing grokking in absence of regularization~\citep{kumar2023grokking}. The importance of the rate of feature learning and the number of necessary features is favored in explanations, challenging the role of the weight norm~\citep{kumar2023grokking}.

Interestingly, a study hypothesizes that memorization constitutes a phase of grokking~\citep{nanda2023progress}. The study finds that grokking includes three distinct stages: memorization, circuit formation, and memorization cleanup~\citep{nanda2023progress}. The study identifies an algorithm that utilizes Discrete Fourier Transforms and trigonometric identities to achieve modular addition through analyzing the model's weights. The circuits enabling this algorithm seem to evolve in a steady manner instead of randomly walking. However, our understanding towards the relationship between memorization and grokking is still limited.

\section{How to Make Use of The Insights?}
In the preceding three sections, we have explored how knowledge is architecturally composed within LLMs (Section 2), and how this knowledge is encoded in their representations (Section 3). Building on these insights, this section emphasizes on how we can leverage our in-depth understanding of LLMs to enhance their performance through editing, improve their efficiency via pruning, and better align them with human values and preferences.

\subsection{Model Editing for Better Performance}
Research has shown that it is possible to edit factual knowledge by modifying the weights of specific neurons in MLPs. One study successfully adopts this approach by altering neural computations related to recall of factual knowledge~\citep{meng2022locating}. Another study expands this method further to allow multiple edits at the same time~\citep{meng2022mass}. Although these methods are effective for targeted edits, their capabilities on updating relevant knowledge and preventing forgetting still require further investigation~\citep{cohen2023evaluating}.

Interestingly, a recent study indicates that the paragraphs memorized by a model can be pinpointed using high-gradient weights in attention heads of the lower layers ~\citep{stoehr2024localizing}. This research employs localization techniques to identify specific attention heads, which are then fine-tuned to unlearn the memorized knowledge. This approach holds promise in enhancing privacy protection in large language models, although a comprehensive evaluation is still needed. 

Besides, facts are also encoded in the representation space, making representation a natural candidate to edit models' outputs. So far, most studies focus on modifying representations at inference time, while the influence of permanent modifications has barely been studied.
A recent work provides a more precise way to edit model representations to change their output distributions~\citep{hernandez2023inspecting}. Instead of only adding the derived vectors into output representations, this study directly changes the embedding of a related entity so as to trigger targeted outputs. As a result, the position of the modified entity in the embedding space has changed, leading to causal influence on model generations.

\subsection{Model Pruning for Better Efficiency}
Causal tracing or causal mediation analysis from mechanistic interpretability serves as one of the fundamental techniques for studying neurons and attention heads. One study uses above method to reveal how the model processes the inputs, showing that the attention mechanism helps models extract query information into the final token in early layers, then result-related information will be incorporated into residual stream in the late MLP layers~\citep{stolfo2023mechanistic}. This finding is meaningful for both pruning and fine-tuning when targeting specific queries.

In contrast to deciphering the inner workings of models, one study examines the differences between pre-training and fine-tuning phases with mechanistic interpretability tools. It reveals that fine-tuning retains all the capabilities learned in the pre-training phase. Transformations between pre-training and fine-tuning stem from ``wrappers'' in MLPs learned on top of models. Interestingly, these wrappers can be eliminated by pruning a few neurons or retraining on an unrelated downstream task~\citep{jain2023mechanistically}. This discovery sheds light on potential safety concerns associated with current alignment approaches.

Different from pruning neurons, the idea of representation engineering, that is directly manipulating representations without the need for optimization or additional labeled data, has also been demonstrated effective in model pruning. Some work attempts to fine-tune models with representation engineering and achieves a comparable and even better performance than state-of-the-art fine-tuning techniques~\citep{wu2024advancing, wu2024reft}.
One work employs forward passes from two topics and derives their difference vectors, which are used in inference time without additional fine-tuning~\citep{turner2023activation}.~\citet{wu2024advancing} also demonstrates the feasibility of fine-tuning models through editing representations. Unlike conventional parameter-efficient fine-tuning (PEFT), representation editing focuses on learning an additional group of trainable parameters to modify representations directly other than models' parameters. And the trainable parameters have been reduced to a factor of 32 compared to that of LoRA~\citep{hu2021lora, wu2024advancing}. Another approach utilizes the distributed alignment search of~\citet{geiger2024finding} to find a set of linear subspace implementing interventions. This method outperforms most PEFT models on a range of tasks~\citep{wu2024reft}.

\subsection{Model Alignment to Human Values}

From the mechanistic perspective, practical applications tend to evaluate model alignments with different tools.
Inspired by induction heads, a recent work measures bias scores of attention heads in pre-trained LLMs, focusing on specific stereotypes. It implemented a method to ensure the accuracy of identifying biased heads by comparing the changes of attention score between biased and regular heads. Through masking identified biased heads, the study effectively reduces the gender bias encoded in the model~\citep{yang2023bias}.
Besides, another work localizes attention heads that are responsible to lie with linear probing and activation patching. A set of intentionally designed prompts is used to instruct LLMs to be dishonest. Meanwhile, linear probes are trained to classify true/false activations of heads. Then, the selected activations are patched with those of honest behaviors to observe the changes of outputs. Multiple attention heads across five layers are causally located in~\citet{campbell2023localizing}.

Existing probing-based bias measurements rely heavily on carefully designed prompts known as prompt engineering~\citep{tamkin2023evaluating}. The effectiveness of these measurements is determined by the comprehensiveness of these prompts. However, the prompts are capable of capturing only recognized biases using a finite set of examples. This fails to provide an inclusive way to uncover biases that have been learned but are not explicitly known. Recently, representation engineering has emerged as a promising avenue for detecting such biases within embedding space.
A notable study suggests that MLPs operate on token representations to alter the distribution of output vocabulary~\citep{geva2022transformer}. After reverse engineering MLPs, it is believed that the output from each feed-forward layer can be seen as sub-updates to output vocabulary distributions, essentially promoting certain high-level concepts. This insight has been used effectively to mitigate toxicity levels in LLMs~\citep{geva2022transformer}.
Another line of work finds multiple representation vectors within MLPs that encourage models' undesired behaviors. These vectors are decomposed using singular value decomposition, allowing researchers to pinpoint specific dimensions that contribute to toxicity~\citep{lee2024mechanistic}.

\section{Conclusions and Looking Beyond}
In this paper, we explore techniques to uncover the inner workings of LLMs through an explainability lens. We focus on two major paradigms of explainability: mechanistic interpretability and representation engineering. We provide a systematic overview of how these techniques can reveal the architectural composition of knowledge within LLMs and the encoding of knowledge in their internal representations. Furthermore, we inspect training dynamics through a mechanistic perspective to explain phenomena like ``grokking'' that can explain generalization abilities of LLMs. Lastly, we reviewed how insights from these explainability analyses can enhance LLM performance through model editing, improve efficiency via pruning, and better align models with human preferences. 

Although there is some preliminary progress in uncovering the inner workings of LLMs, looking beyond, there exist several critical challenges and opportunities. First, LLMs have encoded a vast amount of real-world knowledge into their architectures and parameters. However, 
current research has only revealed a small fraction of the encoded knowledge. Future efforts should focus on developing scalable techniques that can effectively analyze and interpret the intricate knowledge structures embedded within LLMs. Second, 
LLMs have demonstrated remarkable reasoning abilities exhibiting human-like cognitive abilities. However, 
our current understanding of how these high-level reasoning abilities emerge from the interplay of architectural components and training dynamics is limited. More efforts are needed to reveal the intricate mechanisms that give rise to these advanced reasoning capabilities. Third, although the insights gained from mechanistic interpretability and representation engineering have enabled preliminary efforts in areas such as model editing, pruning, and alignment, the progress achieved thus far has been relatively modest. 
More work is required to fully leverage these insights and develop techniques that can substantially improve LLM performance.

\clearpage

\section*{Limitations}
In this paper, we intend to integrate available techniques that enable us to learn the inner workings of LLMs. Despite the valuable perspectives provided, our study has several limitations. First, we do not explore the complete landscape of relevant XAI methods for understanding LLMs, due to space constraints. Other techniques like concept-based explanations, example-based explanations, and counterfactual explanations may also provide some useful insights into the inner workings of LLMs. These methods could potentially uncover additional aspects or offer complementary viewpoints that are not covered by the mechanistic interpretability and representation engineering approaches discussed in this paper. Furthermore, while we try to provide a comprehensive overview of the current state-of-the-art, the field of explainable AI for LLMs is rapidly evolving. New techniques, theories, and findings may emerge that could reshape or extend our understanding of how LLM works. Continuous monitoring and incorporating these developments will be crucial to maintaining a comprehensive and up-to-date perspective on this topic.

\bibliography{custom,example_paper}
\bibliographystyle{acl_natbib}

\clearpage
\appendix

\section{Mechanistic Interpretability}\label{sec:appendix-mi}
Mechanistic interpretability refers to the process of zooming into neural networks to understand the underlying components and mechanisms that drive their behaviors, also known as reverse engineering~\citep{olah2020zoom}. Just as the microscope revealed the world of cells, looking inside neural networks provides a glimpse into rich inner structures of models. This approach diverges from conventional interpretability methods that aim to explain the overall behaviors through features, neural activations, data instances etc. Instead, it draws inspiration from other fields, such as neuroscience and biology, to investigate individual neurons and their connections. By tracking each neuron and weight, an intricate picture emerges on how neural networks operate through interconnected ``circuits'' that implement meaningful algorithms.
On this delicate scale, neural networks become approachable systems rather than black boxes. Neurons play an understandable role and their circuits of connections implement factual relationships about the world. We can thus observe the step-by-step construction of high-level concepts, such as circle detectors, animal faces, cars, and logical operations~\citep{olah2020zoom}. In essence, zooming into the micro-level mechanics of LLMs enables deeper comprehension of their macro-level behaviors. Such mechanistic perspective represents a paradigm shift in interpretability towards unpacking the causal factors that drive model outputs.

\subsection{Role in the General XAI Field}\label{sec:xai-role}
Mechanistic interpretability in XAI represents a paradigm shift towards a deeper and more fundamental understanding of deep neural network (DNN) models~\citep{zhao2023explainability}.

\begin{itemize}[leftmargin=*]\setlength\itemsep{-0.3em}
\item \textbf{\textit{Global} versus \textit{Local} Interpretation:} Mechanistic interpretability diverges from the traditional local focus of XAI, which concentrates on explaining specific predictions made by deep learning models, e.g., feature attribution techniques. Instead, it adopts a global approach, aiming to comprehend DNN models as a whole through the lens of high-level concepts and circuits.

\item \textbf{\textit{Post-hoc} Analysis versus \textit{Intrinsic} Design:} Mechanistic interpretability aims to decipher the complexities inherent in pre-trained DNN models in a post-hoc way. This contrasts with efforts to create models that are mechanistically interpretable by design~\citep{friedman2023learning}.

\item \textbf{\textit{Model-Specific} versus \textit{Model-Agnostic}:} Unlike some XAI methods such as LIME~\citep{ribeiro2016should} and SHAP~\citep{lundberg2017unified}, which are model-agnostic, mechanistic interpretability is a model-specific explanation. It requires tailor-made designs for each distinct LLM, analyzing their unique characteristics.

\item \textbf{\textit{White-box} versus \textit{Black-box}:} Mechanistic interpretability aligns with white-box analysis, requiring direct access to a model's internal parameters and activations. This is in contrast to black-box XAI tools such as LIME and SHAP, which operate solely based on the model's inputs and outputs. 
\end{itemize}
In summary, mechanistic interpretability in XAI is a critical approach to gain a profound understanding of DNN models. It emphasizes a \textbf{global} and \textbf{post-hoc} perspective, focusing on \textbf{model-specific, white-box} analysis to decipher the inner workings and intrinsic logic of complex AI systems. This approach is pivotal to advance transparency and build trust for LLMs, especially in high-stake scenarios where grasping ``why'' behind AI systems is as crucial as the decisions themselves.

\subsection{Why Mechanistic Interpretability?}\label{sec:why-mi}
The question naturally arises: \emph{Why has XAI research on LLMs moved towards the more specialized domain in mechanistic interpretability}? Exploring this shift can shed light on the evolving needs and challenges in this field. In this section, we attempt to look through several factors that we believe have played a major role in steering the shift.

\vspace{2pt}
\noindent\textbf{Alignment Requirement.}\,
In the age of LLMs, the standards for model performance have become more rigorous, not just in terms of accuracy but also in addressing crucial social concerns like dishonesty and fairness. Under this circumstance, the challenge of aligning LLMs with our values and expectations has become a pressing concern, one that demands a deep understanding and effective control of these models. To tackle these challenges, mechanistic interpretability stands out as a promising approach, offering a way to understand the underlying workings of these models.

\vspace{2pt}
\noindent\textbf{Understanding Reasoning Capability.}\,
The field of XAI in machine learning has made significant progress with techniques designed to provide valuable insights to end users, such as feature attributions~\citep{ribeiro2016should} and example-based explanations~\citep{koh2017understanding}. These techniques have been proven to be quite effective in computer vision tasks, where the demands for complex alignment were less strict.
However, as LLMs become more sophisticated, their reasoning capability has transformed from mere pattern recognition to a form of complex, human-like cognition. 
This advancement in LLMs' reasoning abilities renders traditional XAI methods obsolete and less competent in interpreting their behaviors.

\vspace{2pt}
\noindent\textbf{Understanding Inner Working of LLMs.}\,
Moreover, alongside the strong reasoning abilities of LLMs, their notorious deep and intricate architectures are raising new concerns. Since the inner workings of these models are multifaceted and intricate, new challenges in explaining models at the structure level have emerged. Conventional global interpretability techniques, which are adept at uncovering the high-level knowledge acquired in different components of models, fall short when providing sights into the functions and the evolution of knowledge within these models.
This issue is further confounded as LLMs scale aggressively, making neuron-level and layer-level insights increasingly insufficient. This complexity highlights the urgent need for innovative approaches that enable us to zoom in models and provide more in-depth, mechanistic understandings at various levels.

Alternatively, mechanistic interpretability aims to unravel the inner workings of LLMs, providing insights into the ``how'' and ``why'' behind their decision-making processes. 
Specifically, mechanistic interpretability focuses on the causal relationships and underlying mechanisms within models. This not only is more suited to the advanced nature of LLMs, but is also crucial to ensure transparency, trust, and reliability in their applications.

\subsection{Mechanistic Interpretability Theories}\label{sec:mi-theory}
Most of the current work on mechanistic interpretability is based on vision models, and some recent work has begun to investigate Transformer models. In this section, we introduce some core concepts and pivotal phenomenons in the field of mechanistic interpretability. Since LLMs are too complicated to analyze locally, simple yet artificial models are purposely designed to investigate their characteristics and internal mechanisms. We will introduce the main assumptions and observations made under this setting, including \textit{circuits}, \textit{induction heads}, \textit{superposition}, \textit{polysemanticity}, and \textit{monosemanticity}.

\section{Mechanistic Interpretability v.s. Representation Engineering}
In this section, we provide further discussion on different explanation scales of two techniques. Further, we provide our understanding towards their 

\vspace{2pt}
\noindent\textbf{Explanability Scale.}\, These two techniques explain LLMs at opposite scales.

\begin{itemize}[leftmargin=*]\setlength\itemsep{-0.3em}
\item \textbf{Micro-scale}: Mechanistic interpretability focuses on dissecting the intricate inner workings of LLMs at the neuron and circuit levels. It aims at illustrating how models function and process certain tasks with subnetworks.
\item \textbf{Macro-scale}: Representation engineering places representations, rather than neurons or circuits, as the central unit of analysis. The goal is to understand and control cognitive behaviors by studying their manifestations in learned representation spaces. 
\end{itemize}

\vspace{2pt}
\noindent\textbf{Roles in XAI.}\, 
Two techniques are providing multifaceted perspectives in the field of XAI. Representation engineering embodies how well embeddings capture the essence of data. Good representations are crucial to making accurate predictions. The visualization of representation can also implicitly demonstrate the quality of learning. On the other hand, through the lens of mechanistic interpretability, we can investigate relations between models' abilities like generalization and training dynamics. Examining the evolution of models from initialization to generalization, we can reveal characteristics of generalization, such as sparsity.  These characteristics could serve as benchmarks for what constitutes ``good learning''. Apart from that, mechanistic interpretability is known to explain individual functional components and potentially improve model performance in the future.

\vspace{2pt}
\noindent\textbf{Potential to Alignment.}\, At the current stage, both techniques have witnessed preliminary applications in LLM alignment. Mechanistic interpretability plays a crucial role in locating knowledge or biases at the level of attention heads, while representation engineering is primarily employed in targeting undesired behaviors at the level of layers. Despite the distinct focus of each approach within models, both have proven effective in identifying biases and highlighting practical steps for improvement. However, they are still incompetent in uncovering rudimentary causes behind these biases.

\section{Research Challenges}
In this section, we outline the research challenges that deserve future efforts from the community.

\subsection{The Validity of Existing Theories}
While theories that attempt to explain the mechanisms behind the capabilities of transformer models are promising, their empirical support is not definitive. For example, understanding induction heads is key to explain transformer models because they are recognized as foundations for in-context learning abilities. However, as highlighted by~\citet{olsson2022context}, defining what exactly an induction head is remains somewhat elusive. Similarly, the proposition of a mathematical framework to explain circuits inside a simplified network opens up an interesting avenue of research. Although \citet{lieberum2023does} conclude that circuit analysis is feasible on LLMs, this theoretical framework has not been thoroughly tested with empirical studies. Besides, these theoretical models rely on idealized assumptions such as superposition and often lack ground truth. This further complicates the task of validating these theories.

\subsection{The Curse of Dimensionality}
Another challenge is that the parameters we can explain are much less than a third of all parameters in LLMs. These explanations focus on components of attention heads, and although dictionary learning helps to partially understand polysemantic neurons, there is still a vast territory that remains unexplored. The rest majority of these model parameters are tied to MLP layers, which are notoriously difficult to fully comprehend~\citep{olsson2022context}. Their compositions are more complicated than those of attention layers, making the analysis process considerably more arduous and perplexing. For instance, \citet{geva2021transformer} believes that the output of MLPs is a composition of memories including textual patterns and output distributions. \citet{meng2022locating} attempt to modify MLPs to edit factual knowledge in LLMs. However, the effectiveness of editing has been put into doubt by another work~\citep{hase2023does}.

\subsection{Evaluation of Concepts and Circuits}
A key challenge in mechanistic interpretability is validating and ensuring the accuracy of proposed conceptual explanations and functional circuits. Unlike straightforward metrics in machine learning to assess predictions, interpretation evaluation lacks clear ground truth. As noted in \citet{chan2022causal}, we are short of tools to measure the degree to which explanations interpret the relevant phenomenon. Existing ad-hoc ablation methods, i.e. standard zero and mean ablations, are neither universal nor scalable. Exploring measurements from various angles, such as causal scrubbing, which involves randomly sampling inputs to patch activations without disturbing the input distribution, could enrich our evaluation dimensions.
Moreover, manual inspections are challenging in identifying circuits within LLMs. Our understanding of automatically discovering these circuits is still developing~\citep{wang2022interpretability}. Heterogeneous mechanistic explanations can be generated in networks trained on simple tasks such as modular additions~\citep{zhong2023clock}. This suggests that even in seemingly simple scenarios, the outcomes of circuit analysis can be uncertain. 
Additionally, different models learned on similar tasks might learn same family of 
circuits, but the precise circuits learned by individual networks are not the same~\citep{chughtai2023neural}.

\subsection{Conflicted Explanations}
There are other observations in understanding observations, such as neural collapse~\citep{papyan2020prevalence}, yet there is a notable gap in understanding how these observations are interconnected. The root causes of these observations often lead to conflicting viewpoints. For example, \citet{gromov2023grokking} suggests that grokking might be triggered by the learning of a new feature. Unfortunately, the leap in generalization could be too subtle to notice without a hierarchical model~\citep{gromov2023grokking}. On the other hand, there is some debate around linking grokking with generalization~\citep{levi2023grokking}. Moreover, a significant limitation of these studies is their focus on arithmetic datasets instead of real-world datasets, which casts doubt on how broadly these findings can be applied. To fully understand the generalization of models and reconcile these conflicting views, a holistic examination of how these observations relate to each other and their impact on training dynamics across models is essential.

\end{document}